\pdfoutput=1

\documentclass[11pt]{article}

\usepackage[preprint]{acl}

\usepackage{times}
\usepackage{latexsym}
\usepackage[T1]{fontenc}

\usepackage[utf8]{inputenc}

\usepackage{microtype}

\usepackage{inconsolata}

\usepackage{graphicx}
\usepackage{multirow}
\usepackage{subcaption}
\usepackage{dblfloatfix}
\usepackage{graphicx}
\usepackage{adjustbox}
\usepackage{amsmath}
\usepackage{booktabs}

\author{
 \textbf{Figarri Keisha\textsuperscript{1}},
 \textbf{Prince Singh\textsuperscript{1}},
 \textbf{Pallavi\textsuperscript{1}},
 \textbf{Dion Fernandes\textsuperscript{1}},
\\
 \textbf{Aravindh Manivannan\textsuperscript{1}},
 \textbf{Ilham Wicaksono\textsuperscript{1}},
 \textbf{Faisal Ahmad\textsuperscript{1}},
 \textbf{Wiem Ben Rim\textsuperscript{1}}
\\
\\
 \textsuperscript{1} University College London
}
\usepackage{xcolor}
\begin{document}
\title{\textit{All for law and law for all:}\\ Adaptive RAG Pipeline for Legal Research}
\maketitle
\begin{abstract}
Retrieval-Augmented Generation (RAG) has transformed how we approach text generation tasks by grounding Large Language Model (LLM) outputs in retrieved knowledge. This capability is especially critical in the legal domain. In this work, we introduce a novel end-to-end RAG pipeline that improves upon previous baselines using three targeted enhancements: (i) a context-aware query translator that disentangles document references from natural-language questions and adapts retrieval depth and response style based on expertise and specificity, (ii) open-source retrieval strategies using SBERT and GTE embeddings that achieve substantial performance gains while remaining cost-efficient, and (iii) a comprehensive evaluation and generation framework that combines RAGAS, BERTScore-F1, and ROUGE-Recall to assess semantic alignment and faithfulness across models and prompt designs. Our results show that carefully designed open-source pipelines can rival proprietary approaches in retrieval quality, while a custom legal-grounded prompt consistently produces more faithful and contextually relevant answers than baseline prompting. Taken together, these contributions demonstrate the potential of task-aware, component-level tuning to deliver legally grounded, reproducible, and cost-effective RAG systems for legal research assistance. 
\end{abstract}

\section{Introduction}

Large Language Models demonstrate strong generative capabilities that make them incredibly versatile and useful in various domains and applications. However, they remain vulnerable to hallucinations, which proves to be especially detrimental in high-stakes fields (e.g law), where factual inaccuracies cause significant financial and reputational damage. Retrieval-Augmented Generation (RAG) mitigates this by grounding responses in domain-specific documents, providing concrete sources for the LLM to reference. Previous work investigates the improvement of RAG models for the legal domain, for instance LegalBenchRAG \cite{pipitone2024legalbenchrag} introduces new datasets and benchmarks to better evaluate models which we leverage for our pipeline.

In this work, we build and optimize a novel end-to-end RAG pipeline that can adapt to the user's needs. For this purpose, we introduce three main components of the pipeline: Query Translation, Information Retrieval and Response Generation. First, we assess the user's expertise level. A layperson asking a question about a social networking collecting their data would be different from a legal professional asking the same question. Once the user expertise is estimated, we tailor our responses while referencing related legal documents in our database. For this purpose, our contributions extend beyond prior work by systematically enhancing each stage of the pipeline and demonstrating how open-source methods can rival proprietary systems. Specifically:

\medskip
\noindent\textbf{Contribution 1 (Context-Aware Query Translator):} 
We design a lightweight query pre-processing module that disentangles document references from natural-language questions and classifies queries by expertise (expert vs.~non-expert) and specificity (vague vs.~verbose). These signals guide \emph{context-aware translation}, adapting both retrieval depth and response style to the user's needs.  

\medskip
\noindent\textbf{Contribution 2 (Open-source retrieval strategies):} 
We demonstrate that open-source embedding models (e.g. SBERT, GTE) combined with file-aware query translation can rival (and in some cases outperform) the Text Embeding 3 Large model used in LegalBenchRAG. Our tailored retrieval achieves substantial gains, improving Recall@K by 30--95\% and Precision@K by $\sim$2.5$\times$ for $K>4$, while remaining cost-efficient and reproducible. 

\medskip
\noindent\textbf{Contribution 3 (Evaluation and response generation):} 
We introduce a comprehensive evaluation suite that combines RAGAS faithfulness and answer relevancy with BERTScore-F1 and ROUGE-Recall to assess semantic alignment, factual grounding, and completeness. Using this framework, we systematically evaluate prompt designs and model choices (e.g., GPT-4o-mini, LLaMA-3-8B), finding that a custom legal-grounded prompt consistently produces more \emph{faithful and contextually relevant outputs} than baseline prompting approaches.
\section{Background}

LegalBenchRAG proposed an information retrieval pipeline, measuring precision and recall of retrieval by experimenting with Naïve and Recursive Text Character Split (RCTS) chunking, using OpenAI's embedding model (text-embedding-3-large) and cosine similarity search with and without reranking using Cohere's \texttt{rerank-english-v3.0} model \cite{guha2023legalbench}. The paper concluded that RCTS performed better, while Cohere's reranking reduced overall retrieval performance. LegalBenchRag is a good benchmark for retrieval performance, utilising a standard, well-labelled corpus and query-answer (QA) pairs.

Recent works like Adaptive-RAG \cite{jeong2024adaptiverag} and HyPa-RAG \cite{kalra2025hyparag} significantly influenced our system design. Adaptive-RAG introduced query-aware reasoning paths by using classified query complexity to guide retrieval depth. This inspired our use of complexity prediction to adapt chunk size and prompt design during response generation. HyPa-RAG's hybrid retrieval and query rewriting approach, employing dense and sparse retrievers guided by query complexity, motivated us to explore adaptive retrieval configurations based on document reference relevance. We extend these ideas with our query translation module, splitting each input into a document reference and main question, enabling more precisely targeted retrieval and informed generation.

From a practical standpoint, adopting high-performing open-source embedding models is more desirable. Therefore, we focus on evaluating multiple open-source models to determine whether any can match or exceed the retrieval performance of LegalBenchRAG. Additionally, while LegalBenchRAG provides strong retrieval benchmarks, it does not explore end-to-end response generation. This leaves an open question: How well does their retrieval method integrate into a complete RAG pipeline incorporating improved query translation and response generation to produce more accurate responses? To investigate this, we addressed three hypotheses:
\begin{enumerate}
    \item \label{itm:hypothesis-1} Can open-source embedding models match or outperform the retrieval performance of proprietary models?
    \item \label{itm:hypothesis-2} Do alternative similarity search methods lead to improved retrieval results?
    \item \label{itm:hypothesis-3} Can reranking with different encoder models enhance retrieval effectiveness?
\end{enumerate}

\section{Data}
\subsection{Dataset}

We use the LegalBenchRAG corpus, which was derived from the LegalBench \cite{guha2023legalbench} dataset, a collaboratively constructed reasoning benchmark consisting of 162 tasks covering six different types of legal reasoning. Legal experts annotated queries by highlighting relevant text in source documents. LegalBench-RAG data has two primary components: the original corpus and the QA pairs. The corpus includes \texttt{.txt} documents from four legal domains: ContractNLI (contract understanding and natural language inference), CUAD (Contract Understanding Atticus Dataset for reviewing legal contracts), MAUD (merger and acquisition understanding dataset), and PrivacyQA (privacy policy question answering). It excludes documents that were not requested/targeted by at least one query in LegalBenchRAG. The QA pairs are directly linked to the documents within the corpus. Each query is associated with a list of relevant snippets from source documents that directly answer the query. For each snippet, the file path, the exact quote, and the precise character indices within the document are provided, ensuring a clear reference to the source, enabling accurate evaluation of retrieval performance.

\subsection{Data Sampling}
LegalBenchRAG samples 194 QA pairs per domain and extracts relevant text files to create a lightweight, balanced subset termed LegalBenchRAG-mini, for experimentation. It minimises the number of text files by selecting unique QA pairs within the smallest possible set. We replicated this process, but exact reproduction was impossible due to an unknown random seed. However, our sampling yielded a similarly sized subset of text files and QA pairs, though not identical (see Table \ref{tab:legalbench-narrow}).

\begin{table}[h]
\centering
\renewcommand{\arraystretch}{1.15}
\footnotesize
\begin{tabular}{lrr}
\toprule
\textbf{Dataset} & \textbf{Corpus} & \textbf{QA} \\
\midrule
Contract (LegalBench)   & 18 & 194 \\
Contract (B.E.R.T)      & 20 & 194 \\
\midrule
CUAD (LegalBench)       & 18 & 194 \\
CUAD (B.E.R.T)          & 17 & 194 \\
\midrule
MAUD (LegalBench)       & 29 & 194 \\
MAUD (B.E.R.T)          & 16 & 194 \\
\midrule
Privacy (LegalBench)    & 7  & 194 \\
Privacy (B.E.R.T)       & 7  & 194 \\
\midrule
\textbf{Total (LegalBench)} & \textbf{72} & \textbf{776} \\
\textbf{Total (B.E.R.T)}    & \textbf{60} & \textbf{776} \\
\bottomrule
\end{tabular}
\caption{Corpus and QA sample sizes across four datasets in LegalBench and B.E.R.T.}
\label{tab:legalbench-narrow}
\end{table}

\section{Experimental Setup}
\label{sec: Experimental_setup}
\subsection{Query Translation}
We observe that many entries in the LegalBench dataset follow a rigid pattern specifically mentioning a reference document along with a query in the format ``Consider document A; query B``. To address this, we build a Simple Extractor (SE) that processes the queries by splitting them, removing stopwords (e.g., "Non-Disclosure Agreement"), and embeding the document reference with a sentence transformer (\texttt{all-MiniLM-L6-v2}) to find the best-matching file via cosine similarity. Matches are scored as 1 for correct, -1 for incorrect, and 0 if the similarity falls below a threshold. With the exception of CUAD (Contract Understanding Atticus Dataset) where file names often diverge from the textual content of queries due to their standardized naming convention, we increase the threshold to 0.55 for CUAD to reduce mismatches, while lower thresholds (0.3–0.38) are sufficient for the other datasets. Table~\ref{tab:se_sentiment_scores} reports SE's performance on the original dataset queries with their respective thresholds.

\begin{table}[h]
\centering
\renewcommand{\arraystretch}{1.15}
\footnotesize
\begin{tabular}{lrrr}
\toprule
\textbf{Dataset (threshold)} & \textbf{-1} & \textbf{0} & \textbf{1} \\
\midrule
ContractNLI (0.3)  & 0 & 15  & 179 \\
CUAD (0.55)        & 0 & 114 & 80  \\
MAUD (0.38)        & 0 & 0   & 194 \\
PrivacyQA (0.3)    & 0 & 0   & 192 \\
\bottomrule
\end{tabular}
\caption{Performance of the Simple Extractor (SE) across datasets, measured by match scores at varying thresholds.}
\label{tab:se_sentiment_scores}
\end{table}

Although SE performs well when queries strictly follow aforementioned rigid pattern, many queries deviate from this format in real usage. To test robustness, we rephrase all queries in the datasets using a few-shot prompt, ensuring that the \emph{document reference} (e.g., "In the Non-Disclosure Agreement between CopAcc and ToP Mentors …") is preserved while rewriting the question into a more conversational style (e.g., "Is it clearly stated that the Receiving Party has no rights …?"). We employ \texttt{Flan-T5 Large} to generate these more natural queries, approximating real-world scenarios. 

Our experiments (see Table~\ref{tab:rephrased-scores} in Appendix~\ref{sec:appendix-ner}) show that SE's accuracy drops on rephrased queries, since they no longer follow the semicolon-delimited structure. However, a Named Entity Recognition (NER)–based approach can effectively extract the document reference, resulting in only a modest increase in -1 scores. For datasets such as CUAD, where file names are less indicative of query content, similarity scores were lower, requiring threshold adjustments to mitigate mismatches. Importantly, generating rephrased variants for all four datasets substantially expanded the LegalBench corpus, enhancing its realism for legal document retrieval experiments.

To further enhance query understanding and adapt downstream processing, we added a feature extraction component that classifies queries based on linguistic complexity and specificity:
\paragraph{Expertise Classification:} We employ the Dale-Chall readability formula to estimate the linguistic complexity of each query for all 4 datasets. It is chosen for its proven ability to assess comprehensibility in formal domains like legal text. Unlike metrics based only on sentence length or syllables, it uses a curated list of familiar words, making it better suited for structured, jargon-heavy text, an observation supported by \citet{han2024readability} for legal texts and  \citet{zheng2017readability} for technical health documents. To distinguish between domain experts and laypersons, we use a threshold: scores below 8.0 are labelled non-expert, and 8.0 or above as expert, thus guiding the response generation module to produce answers that are either technically detailed or simplified for broader accessibility.
\paragraph{Vague vs Verbose Classification:} To adapt the retrieval strategy based on query specificity, we introduced a classification mechanism that labels queries as vague or verbose. This distinction enables the system to dynamically adjust the number of chunks retrieved during grounding: vague queries, which are general or under-specified, benefit from broader retrieval, while verbose queries, often multi-part or over-specified, require more selective, targeted context windows. Following HyPA-RAG \cite{kalra2025hyparag}, we constructed a synthetic dataset by rephrasing LegalBench queries for all 4 datasets with \texttt{Meta-LLaMA-3} to create diverse vague and verbose variants. This dataset was used to train DistilBERT for binary classification. Once classified, the vague or verbose label was passed to the retrieval module, guiding chunking behaviour and context scaling for downstream LLM processing.

\subsection{Information Retrieval}
\paragraph{Embedding Model:} We compare three open-source embedding models: SBERT (\texttt{all-mpnet-base-v2}), Distilled SBERT (\texttt{all-MiniLM-L6-v2}), and GTE-Large (\texttt{thenlper/gte-large}) against LegalBenchRAG's \texttt{text-embedding-3-large}. These models span a range of scales: mpnet-base (110M) offers high accuracy, MiniLM (22M) is lightweight and efficient, and GTE-Large is optimised for retrieval and reranking in RAG pipelines. This selection allows us to evaluate how model size and design influence retrieval performance.
\paragraph{Pipeline:} The sampled corpus was chunked using Naïve and RCTS methods, embedded with three models, and stored as JSON vectors. Queries from the benchmarks were also embedded, and similarity search (cosine and BM25) retrieved the top 50 chunks, referred to as unranked for comparison purposes with reranked chunks. However, they are ordered by similarity scores. A reranker then reordered these based on query similarity to produce reranked chunks. Precision and recall were calculated by comparing unranked and reranked results against ground truth spans across chunking–embedding–similarity combinations and k-values (1–50). Additionally, we tested RetroMAE, a SoTA embedding model from the BERGEN paper \cite{rau2024bergen}, against the best-performing configuration to evaluate its effectiveness in the RAG pipeline.     
\paragraph{Evaluation Approach:} Following LegalBenchRAG, we use Precision@K and Recall@K based on span overlap. Retrieved chunks are compared to benchmark QA ground truth by file and span alignment. Precision is computed as overlap length over chunk length, and recall as overlap over ground truth length. With multiple chunks and answers per query, scores are averaged across 194 QA pairs per domain and overall, over varying K-values (see Figure \ref{fig:all_plots_fais}). After identifying the best model, we extend the evaluation to include a SoTA method and increase K to 300 for direct comparison with LegalBenchRAG. This extended evaluation is run only on the top-performing model due to resource constraints.
    
We explore using text-based evaluation in addition to span-based evaluation as a way to measure sentiment similarity between retrieved chunks and ground truth. Ultimately, we decide to proceed using primarily span-based, and use the text-based evaluation as a sanity check by paying attention to the behaviour of precision and recall across K-values, such as in Figure \ref{fig:precision-recall-combined} in Appendix \ref{sec:appendix-text-span}.

\begin{figure*}[t]
  \includegraphics[width=1\linewidth]{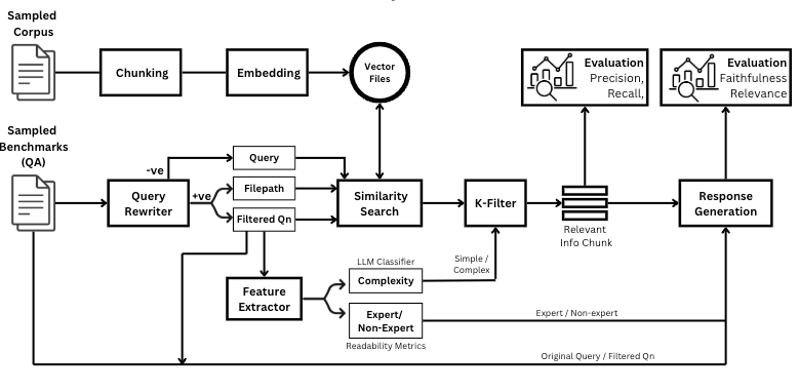} \hfill
  \caption{\textbf{End-to-end pipeline for filtered information retrieval and response generation.} The system begins with a sampled corpus processed via chunking and embedding, stored in vector files. Queries from sampled QA benchmarks are optionally rewritten and analysed for complexity and user expertise. These inputs guide the similarity search and k-filtering process to retrieve relevant information chunks for response generation. Final evaluation is based on both precision/recall and faithfulness/relevance.}\label{fig:pipeline}
\end{figure*}

\subsection{Response Generation and Evaluation (RGE)}
Unlike \cite{pipitone2024legalbenchrag}, which focuses on retrieval, we also explore and evaluate response generation in the legal domain. The RGE is done in two parts: i) evaluation without complexity classifier and readability metrics, to determine the optimal context length, language model, and prompt for the final run and the relevance of each metric for our legal domain use case, and ii) evaluation of the final response employing complexity classifier and readability metrics.

\paragraph{Prompt Designs:} We experiment with several prompts: i) baseline is a standard RAG prompt, ii) zero-shot Chain of Thought (CoT) prompt to assess potential reasoning improvements, and finally iii) a custom-crafted prompt with explicit instructions to enhance accuracy, legal grounding, and relevance.
\paragraph{Evaluation Metrics:} RAGAS (RAG Assessment) is used, especially Answer Relevancy and Faithfulness \cite{ragasgithub2025}. The original RAGAS paper proposed Context Relevance, but it was later deemed unhelpful by the authors \cite{es2023ragas}. Instead, we focus on two RAGAS reference-free metrics: Answer Relevancy, which compares LLM-generated questions (from the model's response) to the original question using similarity scores, and Faithfulness, which checks if the retrieved context supports LLM-generated claims from the response. In addition, we consider using two reference-based metrics: BERTScore-F1 (with LegalBERT) to measure semantic similarity between generated answers and contexts, and ROUGE-Recall to assess completeness through n-gram overlap, to support faithfulness assessment.
\paragraph{Models:} GPT-4o-mini, and LLAMA-3-8B. 
\paragraph{Response generation framework:} Before fitting to the final pipeline and using query translation outputs, we want to find the optimal combination of prompt, model, and k-value by generating responses for all possible combinations using the chunks retrieved by the best model. K-values were varied (1, 3, 5, 10) to compare its effect on generation performance.

\section{Final Pipeline}

The final end-to-end RAG pipeline combined query translation, information retrieval, and response generation stages with parameters that achieved the optimal performance during experimentation (see Figure \ref{fig:pipeline}). Query translation includes query rewriting, file path extractor and complexity classification. Any detected file paths with scores above the dataset-specific thresholds narrow the search to retrieve chunks only from that specific file. Otherwise, the entire database is considered for document retrieval. The feature extraction component classifies the expertise level of the query and adapts the LLM-generated response to match the predicted knowledge level of the user. The query is also classified to be either vague or verbose, tailoring the amount of information retrieved at the retrieval stage. The queries, along with this predicted metadata are passed to the information retrieval stage. At this stage, the sampled corpus is chunked (RCTS) and embedded (SBert). Cosine similarity is then used to retrieve the top-k most relevant chunks for each query. The top-k value is adjusted based on the query type (i.e. vague or verbose). The relevant chunks are then passed on to the response generation module, using a fine-tuned prompt, GPT model and relevant chunks to generate a response, adapted to Expert vs. Non-Expert classification conducted at the query translation stage. The generated response is then evaluated using Faithfulness, answer relevancy, BERTScore-F1 and ROUGE-Recall.

\section{Results and Analysis}

\subsection{Retrieval}
\begin{figure}[t]
\centering
\includegraphics[width=\columnwidth]{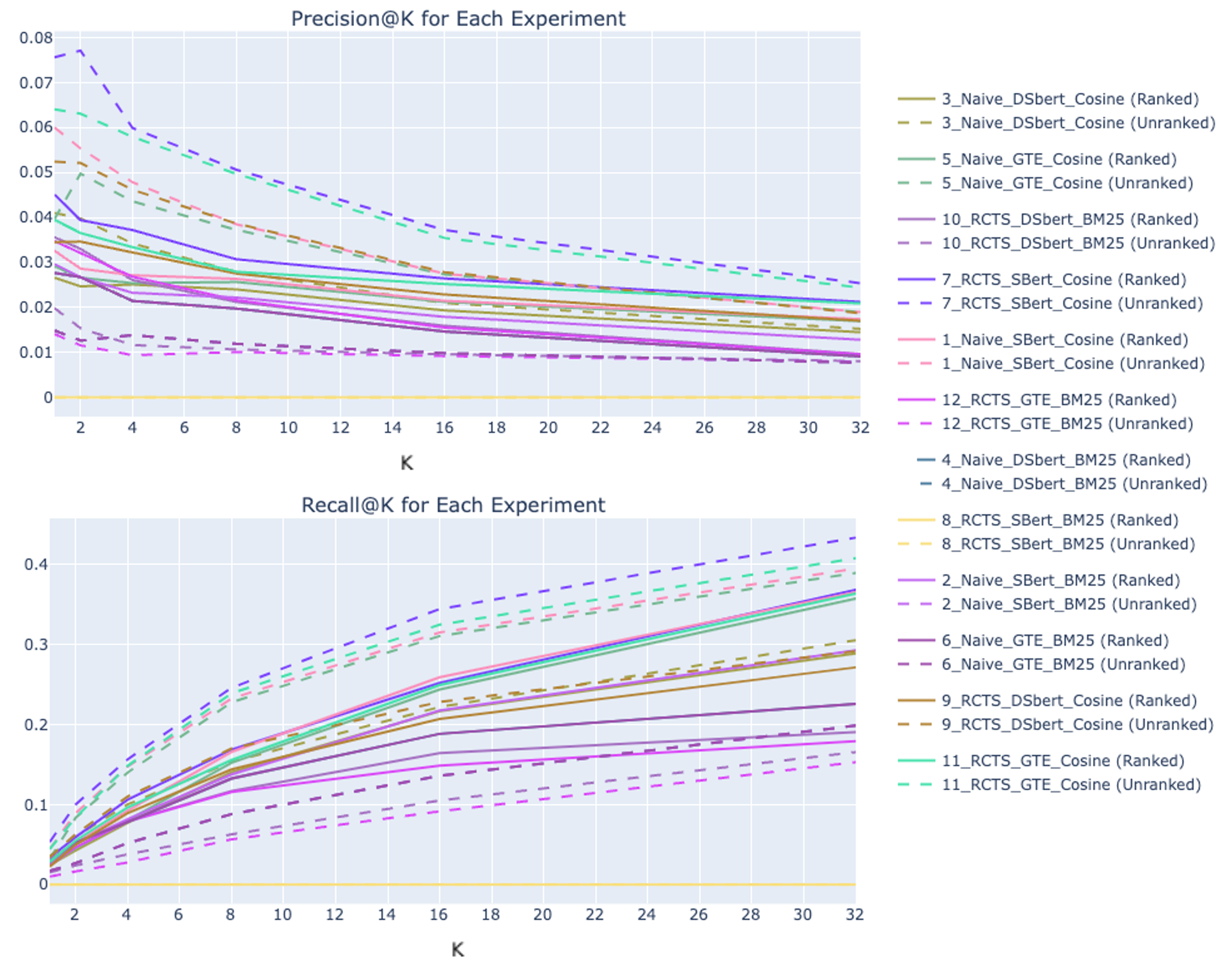}
\caption{\textbf{Precision@K and Recall@K across ranked and unranked experiments.} 
Each curve corresponds to a retrieval configuration (chunking method, embedding model, and similarity search). Precision@K decreases as $K$ increases, while Recall@K improves, reflecting the trade-off between retrieving broader context and maintaining accuracy.}
\label{fig:all_plots_fais}
\end{figure}

The information retrieval performance was evaluated based on the experiments using various models and with the use of Query Translation, comparing against LegalBenchRAG benchmarks as well as SoTA methods.

\begin{figure}[t]
\centering
\includegraphics[width=\columnwidth]{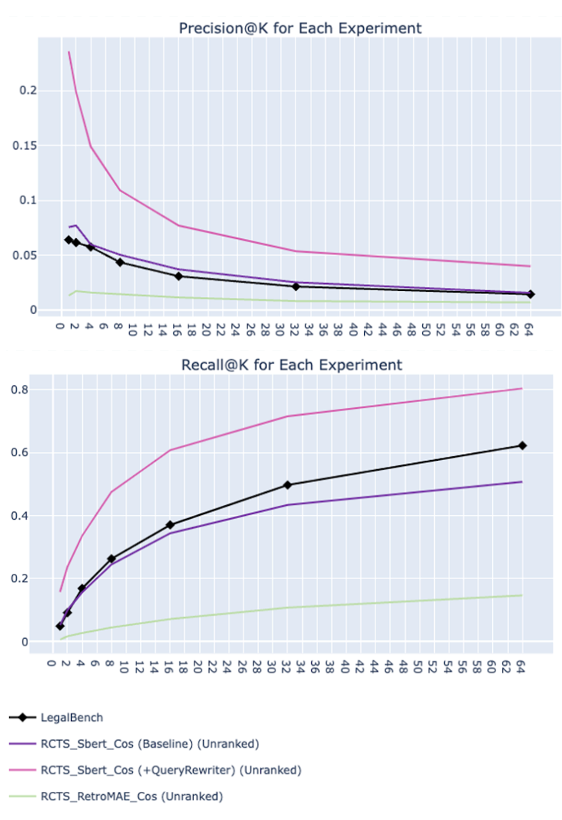}
\caption{\textbf{Precision@K and Recall@K for selected retrieval configurations.} 
Comparison of the RCTS\_SBert\_Cos baseline against its variant with Query-Rewriter, the RCTS\_RetroMAE\_Cos model, and the LegalBenchRAG reference. The plots illustrate how query rewriting and embedding choice impact retrieval quality across different values of $K$.}
\label{fig:baseline_vs_queryrew}
\end{figure}

We observe a few similar findings with LegalBenchRAG. Firstly, RCTS performs better than Naive chunking, although not consistently for all model combinations. Some experiments, such as Naive+SBERT+Cosine and Naive+GTE+Cosine, are in the top 5 performers, at times performing well for individual domains. However, considering overall performance, the top 2 model combinations use RCTS chunking. Thus, overall we can conclude that RCTS outperforms Naive chunking. Secondly, unranked results perform better than reranked ones for cosine similarity and conversely for BM25 similarity, proving our Hypothesis~\ref{itm:hypothesis-3} inconclusive and dependent on model choice. LegalBenchRAG also observes that unranked performs better with cosine similarity. Figure \ref{fig:all_plots_fais} shows these comparisons. Figure \ref{fig:all_plots_fais} also shows that cosine similarity performs better than BM25 similarity search, answering Hypothesis~\ref{itm:hypothesis-2}. The performances may vary for each domain, which is presented in Figure~\ref{fig:precision-recall-k} in Appendix~\ref{sec:appendix-individual-domains}.

\begin{figure*}[!htb]
    \centering
    \includegraphics[width=0.95\textwidth]{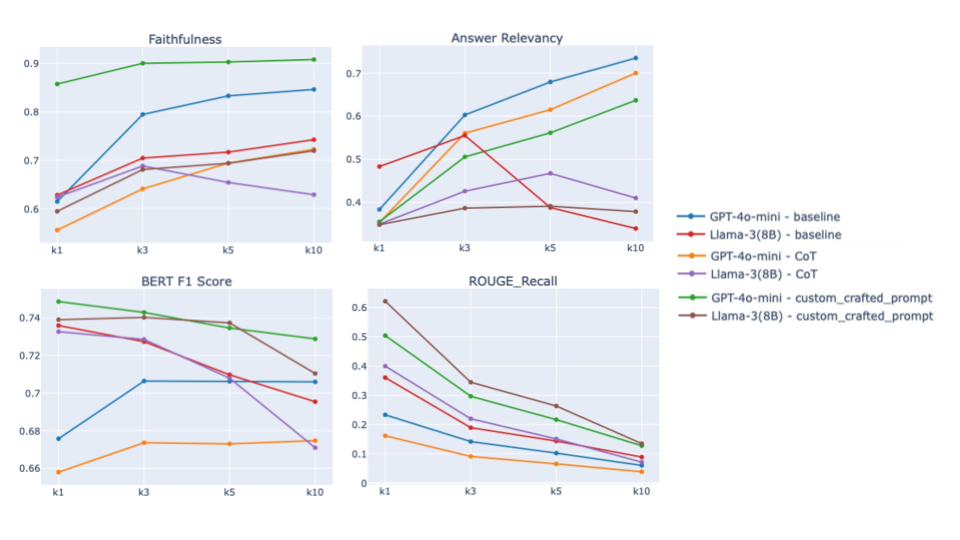}
    \caption{\textbf{Evaluation metrics across prompt strategies and models.} 
    Comparison of GPT-4o-mini and LLaMA-3-8B under three prompting strategies (baseline, zero-shot CoT, and custom legal-grounded). 
    Metrics include faithfulness, BERTScore-F1, ROUGE-Recall, and answer relevancy, showing how prompt design and model choice jointly affect response quality in the legal domain.}
    \label{fig:prompt-eval-metrics}
\end{figure*}

RCTS+SBert+Cosine+unranked is the best-performing model combination considering both precision and recall, followed by RCTS+GTE+Cosine+unranked. Compared to LegalBenchRAG, which uses OpenAI (text-embedding-3-large), our best model performs slightly better on precision for all k-values and very similar for recall for initial k-values, which gets lower as k-value increases (Figure \ref{fig:baseline_vs_queryrew}). To address our Hypothesis \ref{itm:hypothesis-1}, SBERT is an open-source model that can be used reliably in this pipeline with similar performance to the OpenAI model, without incurring significant computational resources.

The best model combination also outperforms the SoTA model (RetroMAE) for both recall and precision (Figure \ref{fig:baseline_vs_queryrew}), proving that established open-source embedding models are still providing sufficient performance for RAG pipeline as compared to the SoTA model. However, we may need to consider that SoTA models may require further finetuning to achieve optimal performance, which could surpass performances seen in these experiments.

\subsection{Response Generation and Evaluation}
The Faithfulness and BERT-F1 score in Figure \ref{fig:prompt-eval-metrics} show that the custom-crafted prompt with GPT consistently performs better than the other combinations across the K values. The Faithfulness and BERT-F1 score show little variation across K value, not providing any conclusive optimal K. In contrast, answer relevancy seems to favour the baseline prompt with GPT across all K values compared to the custom-crafted prompt. Upon further analysis (as shown in the Appendix~\ref{sec:appendix-ragas}), the answer relevancy metric is not well-suited to identify optimal performance. Our analysis shows that answer relevancy primarily reflects the rate of non-committal answers, compared to the actual similarity score itself, downplaying the latter, which we feel is more relevant to this analysis. A modified answer relevancy metric excluding the non-committal multiplier showed comparable results for both the baseline and custom-crafted prompts, reflecting the true cosine similarities. Thus, we can briefly conclude that the use of answer relevancy is not conclusive in finding the optimal performance. However, further studies are required in future work to assess the best alternative metrics. ROUGE-Recall indicates Llama + custom-crafted prompt to be the best performing; however, its trend contradicts Faithfulness and BERT-F1, in contrary to our expectation, considering that all three metrics assess semantic similarity between generated answers and contexts. BERT-F1 exhibits a similar behaviour to Faithfulness as the number of retrieved chunks k increases, which can be attributed to its use of contextual embeddings. Since BERT-F1 measures semantic overlap using embeddings from a domain-specific language model (in our case, LegalBERT), it captures meaning even when the phrasing differs, a critical feature in legal documents where paraphrasing is common. The deviation of ROUGE-Recall could be due to its reliance on surface-level lexical overlap, which does not account for semantic similarity and penalises valid paraphrasing or abstraction (more details are in Appendix~\ref{sec:appendix-rouge}). Therefore, we do not pursue using ROUGE-Recall for evaluation, and identify GPT with a custom-crafted prompt to be an ideal choice as supported by Faithfulness and BERT-F1.

In considering the optimal K-value, we note that the faithfulness stagnates after K=5 and BERT-F1 stagnates with marginal fluctuations for all K-values. Based on Figure \ref{fig:all_plots_fais}, we can see that having too high a K-value will result in a loss of precision but a gain on recall for information retrieval, thus requiring a fine balance in choosing the optimal K-value to trade-off precision and recall. For these reasons, we choose K=5.

\begin{figure}[t]
\centering
\includegraphics[width=0.5\textwidth]{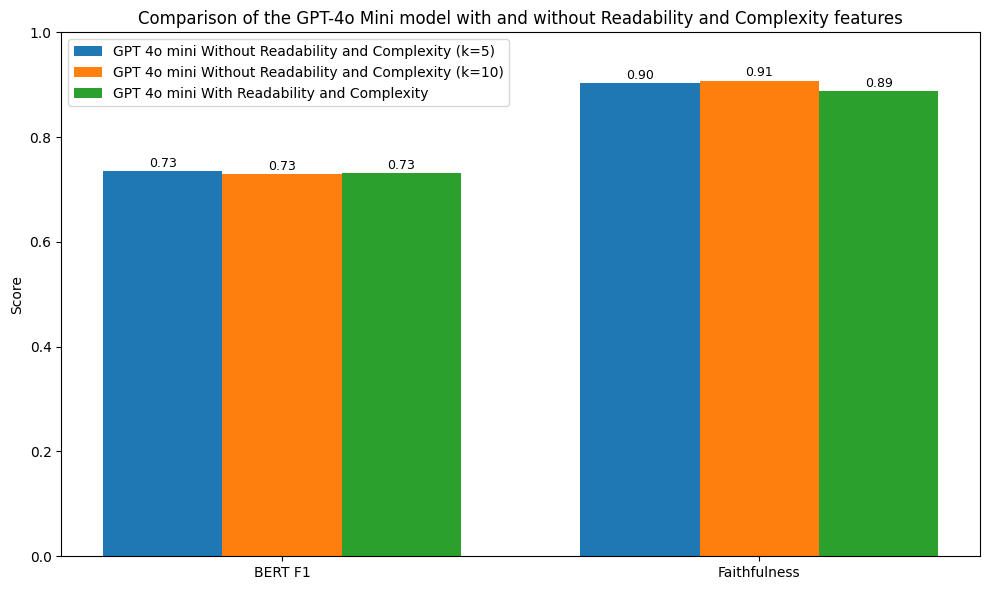} 
\caption{\textbf{Impact of readability and complexity on response generation.} 
Evaluation of responses with and without the readability and complexity classifier shows how adapting output to expert versus non-expert queries affects response style and content, while maintaining overall response quality.}
\label{fig:impact-readability-complexity}
\end{figure}

For the second part of the evaluation, which incorporates the Readability and Complexity classifier (R\&C), we apply K$=$5 for non-expert queries and increase K to 10 for expert queries, to add more details for complex questions. Figure \ref{fig:impact-readability-complexity} shows that including R\&C does not change the performance significantly. 

Qualitative analysis (more details are in Appendix~\ref{sec:appendix-readability}) confirms that the inclusion of R\&C adjusts its tone and content based on query complexity: responses to non-expert queries are concise and free of excessive legal jargon, while responses to expert queries are more detailed and contain strong legal grounding, offering adaptive retrieval strategies without compromising response quality.

\section{Conclusion}

LLMs often suffer from hallucinations, which is a critical issue in the legal domain where obtaining the correct information is crucial. This study addresses this key issue by successfully optimising an end-to-end RAG pipeline for legal documents, advancing beyond the LegalBenchRAG by integrating query translation, retrieval and response generation. 


Our findings highlight the value of adapting retrieval and generation according to query complexity, providing a strong foundation for lightweight legal tools without commercial APIs. \vspace{0.5cm}\\ \indent However, limited resources restricted reranking, fine-tuning, and testing higher K-values compared to LegalBenchRAG. Furthermore, the legal dataset corpus covered only NDAs, M\&A agreements, commercial contracts and privacy policies and did not support queries requiring information from multiple documents. Future work could explore advanced retrievers (e.g. ColBERT, SPLADE) and multi-document queries. Additionally, domain-specific fine-tuning and user experience studies on domain experts vs laypersons could be explored. In general, this research shows a practical and scalable method for implementing RAG pipelines in the legal field, striking a balance between accuracy and accessibility.

\section*{Limitations}

The LegalBenchRAG-mini dataset, while broad, covering NDAs, M\&A agreements, commercial contracts, and privacy policies, is not exhaustive of all types of legal documents. Additionally, each query in this benchmark is answerable by a single document, limiting its ability to evaluate multi-document reasoning. It primarily tests a system's capability to retrieve the correct document and relevant snippets within it.

Our response generation experiments were limited to K=10, and it could still be suboptimal considering that Recall@K continues to improve at larger K values. However, extending response generation to higher K values exceeds the scope and resource limits of our research. These constraints may have affected the generalisability and upper-bound performance of our RAG system, particularly in complex queries requiring broader context. Further work can explore extending the runs beyond these limitations. In-depth analysis of faithfulness and BERT-F1's performance saturation at K=5 should be explored in future research, as such deep meta-analysis of evaluation methods are beyond the scope of this project. Future work can also evaluate the usefulness of the adaptive retrieval performance using a complexity classifier with human-in-the-loop validation.


\bibliography{custom}

\clearpage
\appendix
\section{Named Entity Recognition-based approach}
\label{sec:appendix-ner}

To evaluate robustness on rephrased queries, we compared the Simple Extractor (SE), which relies on semicolon-delimited patterns, against a Named Entity Recognition (NER)–based method. Table~\ref{tab:rephrased-scores} reports the distribution of match scores ($-1$, $0$, $1$) across datasets after rephrasing all queries into more natural forms.

\begin{table}[h]
\centering
\renewcommand{\arraystretch}{1.2}
\footnotesize
\begin{tabular}{l l r r r}
\toprule
\textbf{Model} & \textbf{Dataset} & \textbf{-1} & \textbf{0} & \textbf{1} \\
\midrule
SE  & ContractNLI & 0 & 194 & 0 \\
    & CUAD        & 0 & 194 & 0 \\
    & MAUD        & 0 & 194 & 0 \\
    & PrivacyQA   & 0 & 194 & 0 \\
\midrule
NER & ContractNLI & 4 & 29  & 161 \\
    & CUAD        & 0 & 180 & 14  \\
    & MAUD        & 7 & 24  & 163 \\
    & PrivacyQA   & 0 & 83  & 169 \\
\bottomrule
\end{tabular}
\caption{\textbf{Performance of SE and NER on rephrased queries.} 
Distribution of match scores ($-1$: incorrect, $0$: no match, $1$: correct) across four datasets after rephrasing queries into conversational style. The NER-based approach substantially improves matching accuracy compared to SE, which fails under rephrasing.}
\label{tab:rephrased-scores}
\end{table}

\section{Text vs.~Span-based Evaluation}
\label{sec:appendix-text-span}

\begin{figure}[t]
    \centering
    \begin{subfigure}[t]{\columnwidth}
        \centering
        \includegraphics[width=\linewidth]{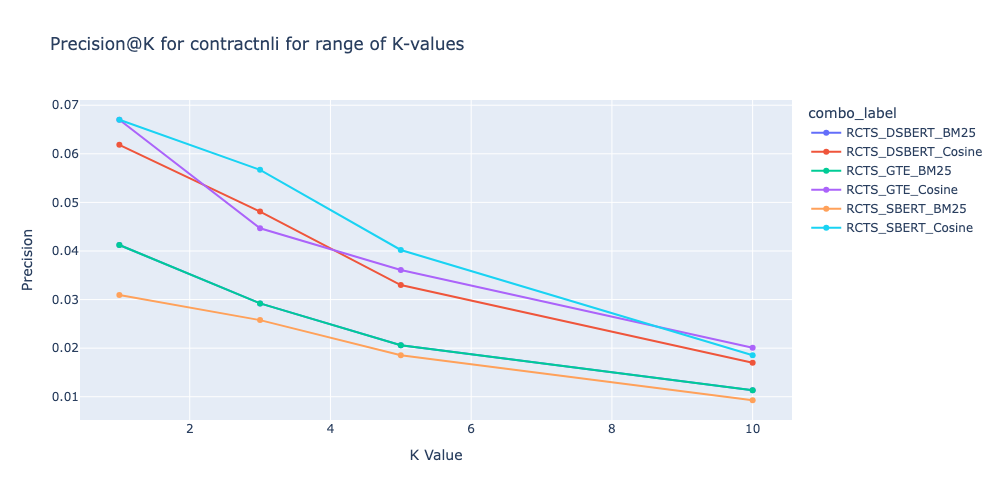}
        \caption{Precision@K for the ContractNLI domain.}
        \label{fig:precision-contractnli}
    \end{subfigure}
    \hfill
    \begin{subfigure}[t]{\columnwidth}
        \centering
        \includegraphics[width=\linewidth]{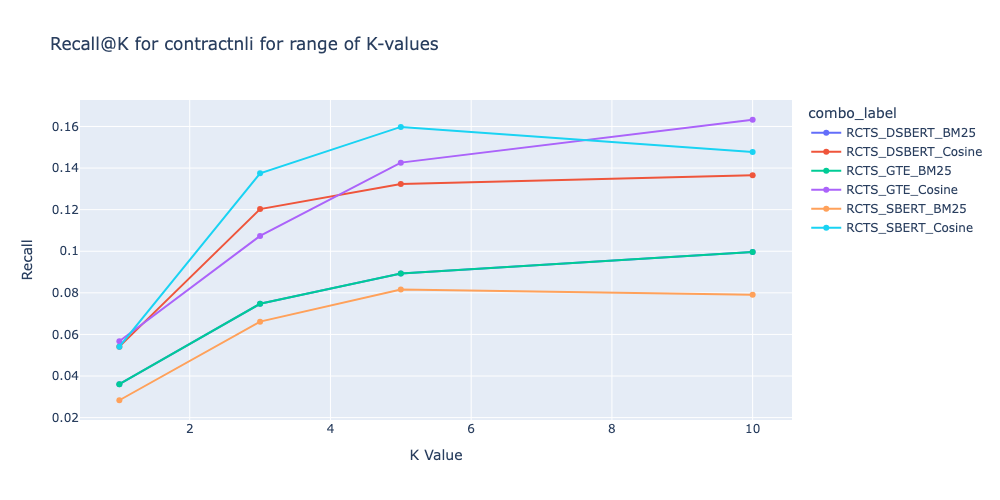}
        \caption{Recall@K for the ContractNLI domain.}
        \label{fig:recall-contractnli}
    \end{subfigure}
    \caption{\textbf{Comparison of text-based and span-based evaluation in ContractNLI.} 
    Precision@K and Recall@K curves show that while span-based evaluation provides stricter alignment with annotated spans, text-based evaluation follows similar trends and was used as a supplementary check.}
    \label{fig:precision-recall-combined}
\end{figure}

To validate retrieval performance, we compared span-based evaluation (matching retrieved text spans to gold annotations) with text-based evaluation (measuring semantic similarity between retrieved chunks and ground truth). While span-based metrics are more reliable for legal retrieval, text-based evaluation served as a sanity check. Figure~\ref{fig:precision-recall-combined} illustrates Precision@K and Recall@K trends for the ContractNLI domain under both approaches.

\section{Information Retrieval Performance by Domain}
\label{sec:appendix-individual-domains}

\begin{figure*}[t]
    \centering
    \includegraphics[width=0.95\textwidth]{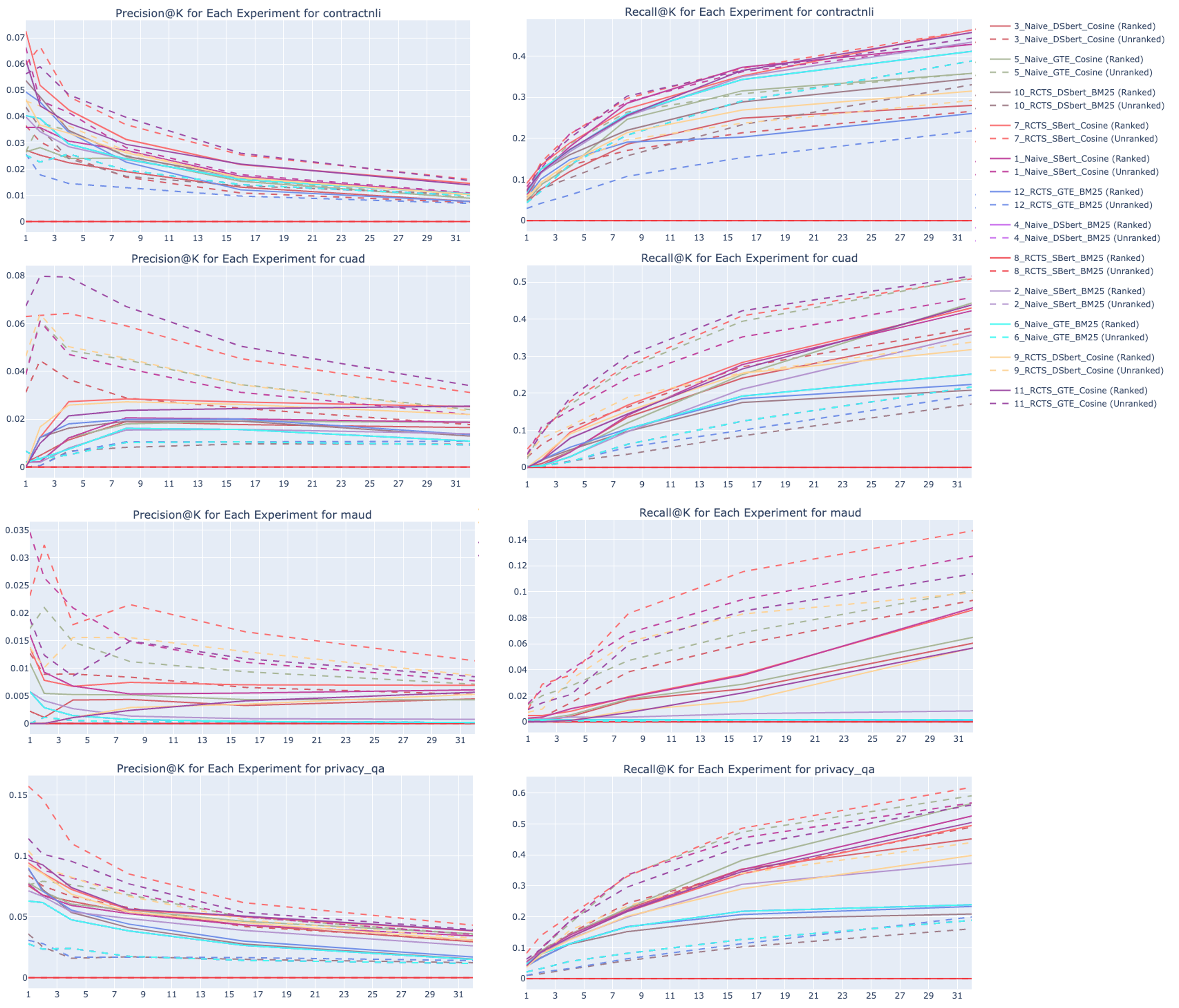} 
    \caption{\textbf{Domain-level Precision@K and Recall@K across retrieval experiments.} 
    Results are shown separately for ContractNLI, CUAD, MAUD, and PrivacyQA. Solid lines denote ranked outputs, while dashed lines denote unranked outputs. The curves highlight variations in retrieval effectiveness across domains and confirm that performance differences are dataset-dependent.}
    \label{fig:precision-recall-k}
\end{figure*}

To provide finer-grained insights, we report domain-specific Precision@K and Recall@K results across all retrieval configurations. Figure~\ref{fig:precision-recall-k} presents the performance curves for ContractNLI, CUAD, MAUD, and PrivacyQA, allowing comparison of ranked versus unranked outputs within each dataset.

\section{Quantitative Look into RAGAS Answer Relevancy} 
\label{sec:appendix-ragas}
On their paper and webpage, answer relevancy is described as a calculation of how relevant the answer is to the original question. This is done by generating a number of artificially generated questions from the response, and calculating the similarity score of them to the original question (3 generated question by default):

\begin{equation}
\text{answer relevancy} = \frac{1}{N} \sum_{i=1}^{N} \frac{E_{g_i} \cdot E_o}{\|E_{g_i}\| \|E_o\|}
\end{equation}

\begin{itemize}
    \item $E_{g_i}$ is the embedding of the generated question $i$.
    \item $E_o$ is the embedding of the original question.
    \item $N$ is the number of generated questions, which is 3 by default.
\end{itemize}

A closer look at their implementation revealed that they also do classification on the answer as a committal and non-committal answer. A non-committal answer is evasive, vague, or ambiguous. For example, "I don't know" or "I'm not sure" are noncommittal answers. If a non-committal answer is detected, the final score will be multiplied by 0. 

We found out that this has a high influence on lowering the total (average) answer relevancy score in a batch of queries compared to the actual similarity score. We finally recalculated answer relevancy for samples in contractNLI document for baseline prompt and human-tuned prompt without non-committal multiplier and found out they both got a very high score around 0.9, and not that different between the two. Analysis of non-committal answers is in the qualitative analysis part.

\begin{figure*}[t]
    \centering
    \includegraphics[width=0.9\textwidth]{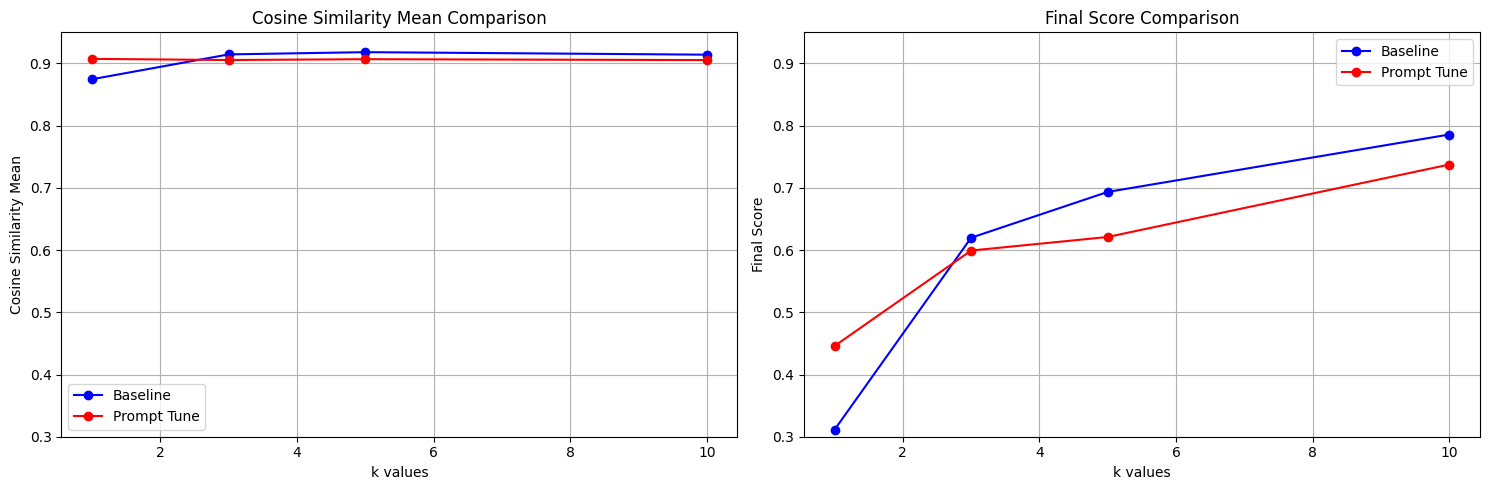}
    \caption{\textbf{Cosine similarity versus final score across $k$ values.} 
    The left plot reports mean cosine similarity excluding the non-committal multiplier, while the right plot shows final scores including the multiplier. Results compare baseline prompts against human-tuned prompts, highlighting how prompt design interacts with scoring criteria.}
    \label{fig:final-score-comparison}
\end{figure*}

We conclude that the answer relevancy score alone is not a definitive indicator of model quality, as it mainly reflects the rate of non-committal answers. Recalculated scores excluding this factor yield similarly high and comparable results across models. Moreover, the non-committal count decreases for both prompts as $k$ grows, suggesting that additional retrieved context contributes to more decisive responses, in line with Recall@K trends in information retrieval.

\begin{figure}[t]
\centering
\includegraphics[width=\columnwidth]{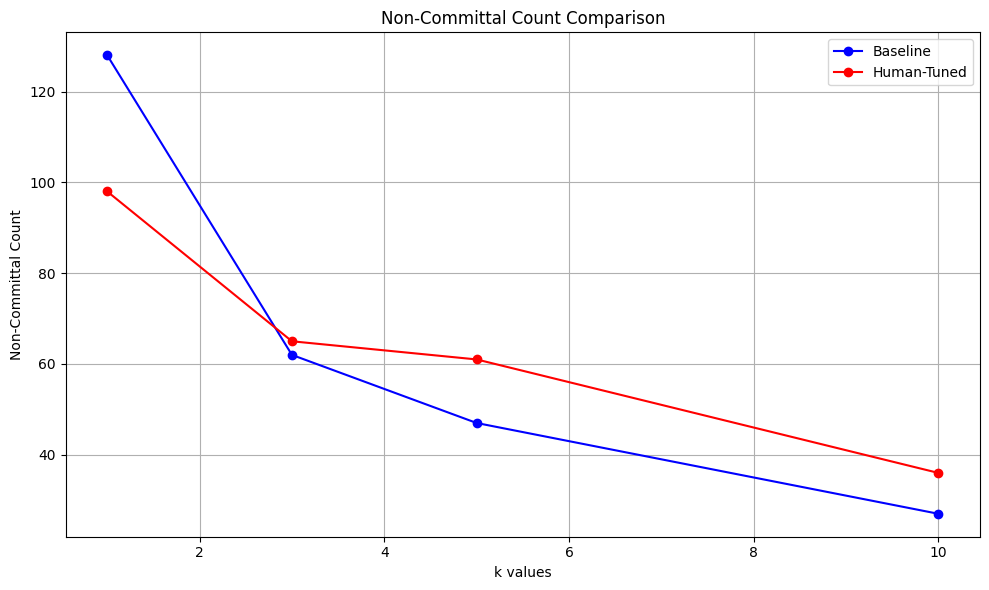} 
\caption{\textbf{Non-committal responses across $k$ values.} 
The frequency of non-committal answers decreases as more context is retrieved, supporting the observation that higher recall contributes to more decisive model outputs.}
\label{fig:non-committal-count}
\end{figure}

Another parameter in answer relevancy is the number of artificial questions generated. Our initial hypothesis was that increasing the number of generated questions would provide an advantage as the number of retrieved contexts ($k$) grows. To test this, we compared the default setting of three generated questions with an extended setting of five, focusing on the ContractNLI dataset with GPT-4o-mini.

\begin{figure}[t]
\centering
\includegraphics[width=\columnwidth]{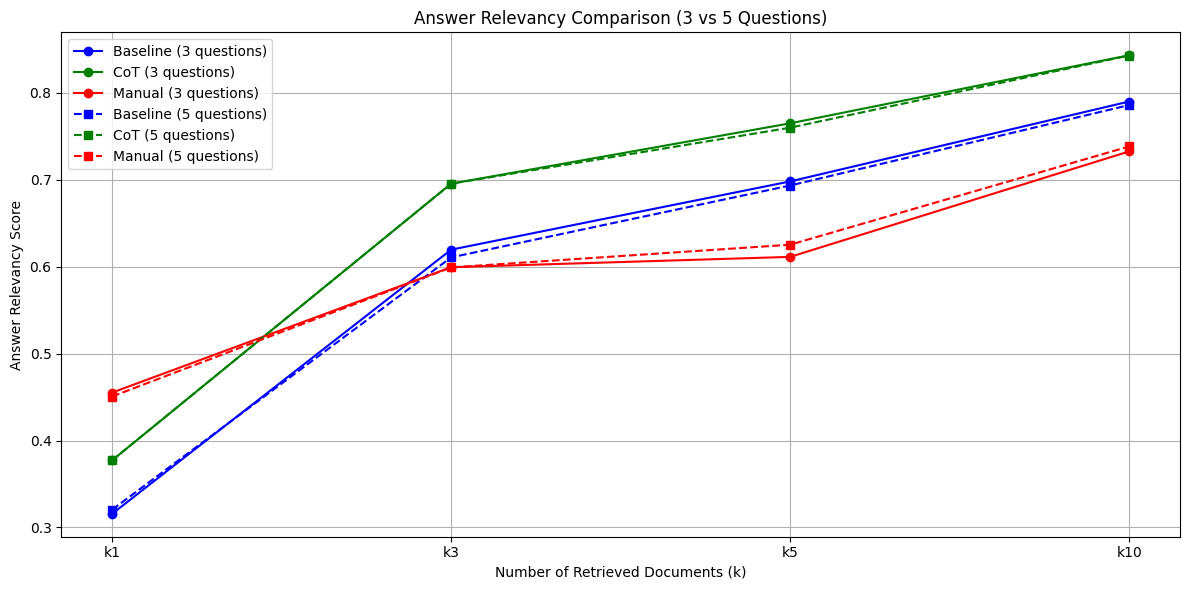} 
\caption{\textbf{Answer relevancy with different numbers of generated questions.} 
Comparison of default (3) versus extended (5) artificial questions on ContractNLI using GPT-4o-mini, showing how the number of generated questions influences answer relevancy as $k$ increases.}
\label{fig:answer-relevancy-questions}
\end{figure}

It can be seen that the difference between scores from 3 questions and 5 questions are minuscule. We conclude to just use the default number of 3 questions for all our evaluations.

\section{Using Rouge as an Evaluation Metric}
\label{sec:appendix-rouge}
Our initial setup included ROUGE recall (average of ROUGE-1, ROUGE-2, and ROUGE-L) as one of the core metrics to evaluate content overlap with reference answers. However, as shown in Figure~\ref{fig:prompt-eval-metrics}, we observe that ROUGE recall exhibits a declining trend as the number of retrieved chunks (k) increases, in contrast to metrics such as BERTScore F1 and RAGA Faithfulness/Relevancy, which stabilise or improve. This discrepancy motivated an in-depth investigation of the limitations of ROUGE in this context.

ROUGE was originally designed to evaluate extractive summarization by computing n-gram overlap between generated and reference texts \cite{chalkidis2020legalbert}. While effective for summarisation and simple question-answering scenarios, it falls short in evaluating abstractive or semantically equivalent generation, especially when the wording differs from the reference but the meaning is preserved.

\begin{itemize}
    \item Semantic vs. Lexical Similarity: ROUGE heavily relies on surface-level lexical overlap, penalizing answers that are semantically correct, but lexically divergent from the gold answer. In contrast, as the number of retrieved chunks (k) increases, the model has more context to paraphrase or synthesise information, often resulting in semantically accurate but lexically novel responses. This leads to low ROUGE scores despite high answer quality.
    \item Recall-Oriented Nature: ROUGE-recall favors longer answers that capture more reference n-grams. However, in a RAG setting with an increase in k, the model may generate more focused and concise responses due to better context resolution. This penalises shorter, yet more precise answers, leading to deceptively low ROUGE recall.
    \item Empirical divergence: As shown in our results, ROUGE recall decreases monotonically with increasing k, whereas RAGAS Faithfulness and Relevancy (which evaluate whether the answer is supported by retrieved knowledge and relevant to the question) increase or stabilise. This indicates that ROUGE recall is misaligned with the actual performance of the system, potentially misleading model development and tuning.
\end{itemize}
Justification for Removal: Given these limitations, we argue that ROUGE recall is an inadequate metric for evaluating RAG-based question answering systems where: (i) Answers are generated using paraphrased or synthesised language. (ii) Multiple retrievals lead to more diverse but valid completions. (iii) Lexical similarity is not a reliable proxy for answer quality.

Hence, for more meaningful and context-aware evaluation, we focus on RAGAS Faithfulness and Relevancy, which accounts for whether the answer is grounded in retrieved content; BERTScore F1 \cite{ragasgithub2025}, which reflects semantic similarity at the embedding level. By excluding ROUGE recall, we avoid penalising correct responses that diverge lexically, and better align our evaluation metrics with the goals of faithful and relevant generation.

\section{Qualitative Analysis of GPT-4o Mini With Readability and Complexity} 
\label{sec:appendix-readability}

From the sample responses, we observe that applying the complexity classifier yields concise answers for simpler queries (e.g., Query 3), making them more accessible, while expert queries receive additional detail (e.g., Queries 1 and 2) to better match the required level of specificity.

\begin{quote}
    Query 1: \\
    Consider the Non-Disclosure Agreement between CopAcc and ToP Mentors; Does the document permit the Receiving Party to retain some Confidential Information even after its return or destruction?

    Classification: Expert (K=10)

    Response With R\&C: The NDA does not permit the Receiving Party (Mentor) to retain any Confidential Information after its return or destruction. According to Clause 6, upon the Organiser's request, the Mentor is required to:
    \begin{itemize}
        \item Discontinue all use of Confidential Information,
        \item Return all materials containing Confidential Information,
        \item Destroy any copies of materials that contain Confidential Information,
        \item Erase or destroy any Confidential Information stored in computer memory or data storage,
        \item Remove Confidential Information from any software or database used by the Mentor.
    \end{itemize}
    Furthermore, the Mentor must provide a written warranty within ten days confirming these actions. This indicates a clear obligation to eliminate any retention of Confidential Information. 

    Response with K=5: The document similarly requires the Mentor to discontinue, return, and destroy all Confidential Information, including computer-stored data, and to warrant in writing that these actions are complete. There are no provisions allowing retention.

    Observation: The R\&C response is more detailed, emphasising the lack of retention rights.
\end{quote}

\begin{quote}
    Query 2: \\
    Consider the Data Use Agreement in New York City; Does the document restrict the use of Confidential Information to the purposes stated in the Agreement?

    Classification: Expert (K=10)

    Response With R\&C: Yes, the Agreement restricts the use of Confidential Information to the purposes stated, specifically those in Attachment B. The Data Recipient is further prohibited from disclosing identifiable information and must obtain written approval before releasing public information, reinforcing these restrictions.

    Response with K=5: Yes, the Agreement limits use to Attachment B purposes, and explicitly requires that any published work must not contain identifiable information.

    Observation: The R\&C response highlights confidentiality obligations more thoroughly.
\end{quote}

\begin{quote}
    Query 3: \\
    Consider the Marketing Affiliate Agreement between Birch First Global Investments Inc. and Mount Knowledge Holdings Inc.; What is the expiration date of this contract?

    Response With R\&C: The Agreement expires on July 22, 2019, as stated: ``the Agreement shall expire and terminate automatically without further notice on July 22, 2019.''

    Response with K=10: The Agreement is valid for five (5) years from its effective date of July 22, 2014, thus expiring on July 22, 2019, unless earlier terminated with 30 days' notice.

    Observation: The R\&C response is more concise, directly stating the expiration date.
\end{quote}

\end{document}